\documentclass{article}

\usepackage{PRIMEarxiv}

\usepackage[utf8]{inputenc} 
\usepackage[T1]{fontenc}    
\usepackage{hyperref}       
\usepackage{url}            
\usepackage{booktabs}       
\usepackage{amsfonts}       
\usepackage{nicefrac}       
\usepackage{microtype}      
\usepackage{lipsum}
\usepackage{fancyhdr}       
\usepackage{graphicx}       

\usepackage{adjustbox}
\usepackage{amsmath}
\usepackage{array}
\usepackage{booktabs}
\usepackage{longtable}
\usepackage{caption}
\usepackage{multirow}
\usepackage{multicol}
\usepackage{rotating}
\usepackage{tabularx}
\usepackage{url}
\usepackage[table,xcdraw]{xcolor}
\usepackage{hyperref} 
\usepackage{balance}

\title{
    Few-Shot Prototype Head Adaptation for On-Device ECG Personalization on PSoC~6
    \thanks{
        The authors would also like to thank the \textit{Fundacão de Amparo à Pesquisa do Estado de Minas Gerais} (FAPEMIG, grants APQ-01768-24), \textit{Coordenação de Aperfeiçoamento de Pessoal de Nível Superior} - Brazil (CAPES) - Finance Code 001, \textit{Conselho Nacional de Desenvolvimento Científico e Tecnológico} and Universidade Federal de Ouro Preto (PROPPI/UFOP) for supporting the development of this study.
    }
}

\author{
    Guilherme A. L. Silva$^1$, Pedro H. L. Silva$^2$, Gladston J. P. Moreira$^2$ and Eduardo J. da S. Luz$^2$\\
    $^1$Computing Department, Universidade Federal de São João del-Rei -- São João del-Rei, Brazil\\
    \texttt{guilherme.silva@ufsj.edu.br}\\
    $^2$ Computing Department, Federal University of Ouro Preto --
    Ouro Preto, Brazil\\
}

\begin{document}
\maketitle

\begin{abstract}
    Wearable and bedside electrocardiogram (ECG) monitors must adapt to patient-specific morphology to maintain arrhythmia detection accuracy across users, yet personalization is typically performed offline and cannot account for individual physiology, electrode placement, or recording drift. On-device adaptation by backpropagation is expensive for microcontroller-class medical devices because it requires an optimizer state, repeated backward passes through convolutional layers, and labeled arrhythmic beats that may not be available at deployment time. This letter proposes prototype-only head adaptation as a compact personalization primitive for TinyML ECG systems. A one-dimensional convolutional neural network (1-D CNN; 1,314 parameters and 72.6k multiply--accumulate operations per beat) is trained offline on the MIT-BIH Arrhythmia Database under an inter-patient protocol, frozen as a feature extractor, and exported to a PSoC~6 microcontroller. Patient-specific adaptation then reduces to computing closed-form class means in a 32-dimensional embedding space, requiring no convolutional backward pass, no iterative optimization, and only one forward pass per support beat. Prototype adaptation improves inter-patient macro-F1 from 0.635/0.639/0.646 to 0.731/0.771/0.797 at 1/5/10-shot, outperforming linear stochastic-gradient-descent (SGD) head fine-tuning at every shot count for the target tiny backbone. On-device replay over 18 one-shot episodes on a PSoC~6 Cortex-M4F matches the host macro-F1 for the prototype head (0.798), with 11.39~ms per beat, 5.2~KB flash, and 22.2~KB SRAM. A restricted variant that updates only the normal-class prototype from passively buffered sinus beats yields a consistent +0.05 macro-F1 gain, reducing the annotation burden during initial deployment.
\end{abstract}


\section{Introduction}
    Patient-specific ECG morphology varies with physiology, electrode placement, and recording conditions, causing models trained on one population to degrade on unseen individuals under inter-patient evaluation protocols~\cite{dechazal2004automatic}. Wearable and bedside arrhythmia monitors, therefore, need a mechanism to personalize a deployed classifier for each new user without re-running a full training pipeline in the cloud. Tiny machine learning (TinyML) has made neural-network inference feasible on microcontroller-class IoT devices~\cite{banbury2021mlperf,lin2020mcunet}, bringing the classifier physically close to the sensor, but the question of how to \emph{adapt} that classifier on the same constrained device remains open.

    Two barriers make on-device ECG personalization hard. The first is \emph{computational}. Backpropagation through convolutional layers requires optimizer state, repeated backward passes, and working memory that exceeds the budget of sub-milliwatt medical microcontrollers. Prior work has shown that on-device training is possible under 256~KB SRAM~\cite{lin2022ondevice} and that 1-D CNN personalization for human activity recognition is feasible on STM32 hardware~\cite{craighero2024ondevice}, yet both approaches rely on gradient-based updates that are costly relative to a medical device's strict power and latency envelope. The second barrier is annotation. Since arrhythmic events are rare, the newly deployed monitor accumulates hundreds of normal sinus beats before a single labeled abnormal beat is available. Gradient-based head fine-tuning cannot begin calibration from normal-only data because it requires labeled examples from every class to update the weight matrix meaningfully. Self-supervised and human-in-the-loop methods have highlighted this gap but remain constrained by limited medical equipment resources~\cite{silva2024bridging}.

    This letter addresses both barriers with a \emph{prototype-only head adaptation} mechanism. A compact 1-D CNN (1,314 parameters, 72.6k multiply–accumulate operations -- MACs -- per beat) is trained offline on the MIT-BIH Arrhythmia Database~\cite{moody2001impact} under the inter-patient protocol~\cite{dechazal2004automatic}, and its convolutional backbone is then frozen and exported to a PSoC~6 microcontroller. Patient-specific adaptation reduces to computing closed-form class means in a 32-dimensional embedding space, one forward pass per support beat, no backward pass, no optimizer state. A restricted variant further reduces the annotation burden by updating only the normal-class prototype from passively buffered sinus beats, inheriting the abnormal prototype from offline training.

    We verify the proposal through three complementary evaluations. (i)~\emph{Accuracy:} prototype adaptation improves inter-patient macro-F1 from 0.635/0.639/0.646 to 0.731/0.771/0.797 at 1/5/10-shot on eligible DS2 patients (vs.\ 0.664 pre-adaptation on all 22 DS2 records), outperforming a linear-SGD head baseline on a tiny backbone neural network. The restricted normal-only variant yields a consistent +0.05 macro-F1 gain without any arrhythmia annotations. (ii)~\emph{On-device fidelity:} aggregate replay over one-shot DS2 episodes on the target Cortex-M4F matches host macro-F1 (0.798 macro-F1, absolute deviation 0.000), confirming that the closed-form update is numerically faithful on fp32 hardware. (iii)~\emph{Embedded cost:} Data Watchpoint and Trace (DWT) cycle counter profiling reports 11.39~ms per beat (2.8\% of the 400~ms RR interval at 150~BPM), with a model-centric footprint of 5.2~KB flash and 22.2~KB SRAM on a device with 1~MB flash and 288~KB SRAM.

\section{Prototype-Only Head Adaptation}

\begin{figure}[!htb]
        \centering
        \includegraphics[width=\columnwidth]{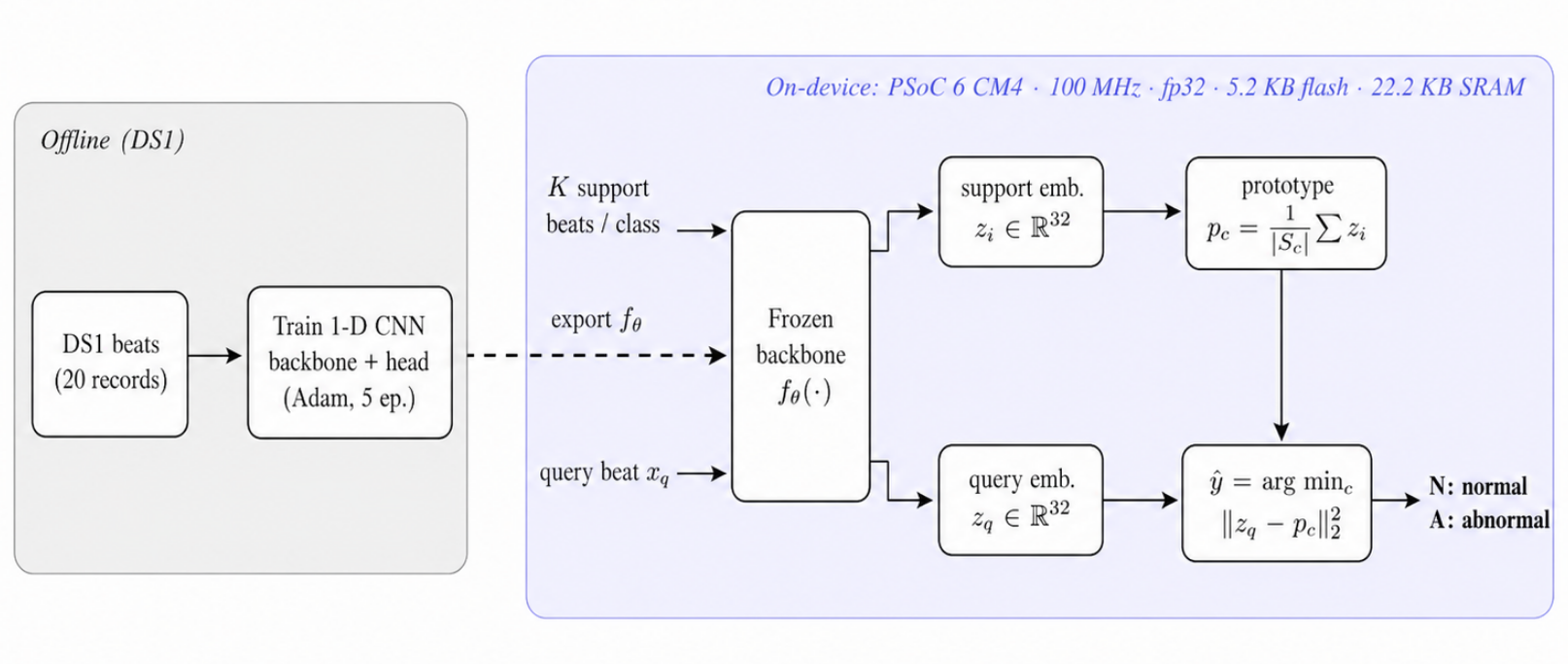}
        \caption{Prototype-only ECG personalisation pipeline. \emph{Offline:} a compact 1-D CNN is trained on DS1 under an inter-patient split; the frozen backbone is exported as native-C arrays. \emph{On-device (PSoC~6):} the backbone maps support beats to 32-D embeddings; per-class prototypes are computed as closed-form means. Query beats are classified by nearest-prototype distance---with no backward pass, no optimizer state, and no backbone writes.}
        \label{fig:pipeline}
    \end{figure}

    Fig.~\ref{fig:pipeline} summarizes the deployment flow. In the offline phase, a compact 1-D CNN is trained on DS1. At export time, the convolutional backbone is frozen and stored as constant $C$ arrays. At deployment, support beats from the target patient are passed through the frozen backbone, and only the classification head is adapted.
    
    Each ECG beat is represented as a 200-sample window centered at an annotated R-peak. The target backbone has two convolutional blocks $(1{\rightarrow}8{\rightarrow}16)$, each with a kernel size of 5, ReLU, max-pooling, global average pooling, and a linear projection to $D=32$. A 2-way head map embedding to normal/abnormal logits. The tiny model has 1,314 parameters and 72,576 MACs per beat; medium and large variants (Table~\ref{tab:scaling}) scale to 2,618/157k and 4,930/274k, respectively. Beat segmentation and R-peak detection are assumed to be provided upstream by the monitoring system; this work focuses on the embedded personalization primitive.

    Let $f_\theta:\mathbb{R}^{200}\rightarrow\mathbb{R}^{D}$ denote the frozen backbone, with $D=32$. In the full few-shot setting, the patient-specific support set is
    \begin{equation}
        S=\{(x_i,y_i)\}_{i=1}^{K\cdot C},
    \end{equation}
    where $x_i\in\mathbb{R}^{200}$, $y_i\in\{0,\ldots,C-1\}$, $C=2$, and $K$ is the number of support beats per class. For class $c$, define
    \begin{equation}
        S_c=\{x_i \mid (x_i,y_i)\in S,\; y_i=c\},
    \end{equation}
    with $|S_c|=K$ in the balanced protocol. Each support beat is embedded as $z_i=f_\theta(x_i)$, and the class prototype is the mean embedding
    \begin{equation}
        p_c=\frac{1}{|S_c|}\sum_{x_i\in S_c} f_\theta(x_i).
    \label{eq:proto}
    \end{equation}

    Although richer prototype estimators have been proposed in recent few-shot learning work, including task-guided diffusion \cite{du2023protodiff} and prototype--neighbor hybrid methods \cite{jiang2025prototype}, these approaches typically rely on meta-training, auxiliary networks, or iterative inference. Such mechanisms are not aligned with the memory and latency constraints of a Cortex-M4F deployment. We, therefore, deliberately use the closed-form class mean, since it is the simplest estimator, requires only support-set inference and vector additions, and remains effective when the frozen backbone yields a compact embedding space.

    A query beat $x_q$ is embedded by the frozen backbone and scored against each prototype using the negative squared Euclidean distance:
    \begin{equation}
        s_c(x_q)=-\left\|f_\theta(x_q)-p_c\right\|_2^2.
    \end{equation}
    The predicted class is
    \begin{equation}
        \hat{y}(x_q)
        =
        \arg\max_{c} s_c(x_q)
        =
        \arg\min_{c}\left\|f_\theta(x_q)-p_c\right\|_2^2.
        \label{eq:decision}
    \end{equation}
        
    This update requires only support-set inference and class-wise additions in a 32-D space. It avoids storing the optimizer state and avoids backward propagation through the convolutional layers. Support beats from each class cluster around their respective centroids in the frozen embedding space, and the resulting prototypes define patient-specific decision boundaries without any gradient steps. We compare it against a linear-SGD head baseline, in which the frozen backbone is held fixed while the final linear layer is updated for 50 SGD steps using cross-entropy.

    The per-class structure of the prototype update enables a clinically important restricted variant. In routine monitoring, arrhythmic beats are rare, and a device may accumulate hundreds of normal sinus beats before any arrhythmic event occurs, making annotation-dependent calibration impractical. We therefore evaluate a setting in which only normal-beat support is used; $p_0$ is updated via Equation~\eqref{eq:proto} while $p_1$ is inherited from DS1 training, enabling passive annotation-free calibration that begins as soon as sufficient normal beats have been buffered.

\section{Protocol and PSoC~6 Implementation}

    Experiments use the MIT-BIH Arrhythmia Database~\cite{moody2001impact} under the inter-patient split of de Chazal \textit{et al.}~\cite{dechazal2004automatic}. Dataset 1 (DS1) is used for training/validation, and dataset 2 (DS2), for held-out testing. The task is binary normal-versus-abnormal classification, where the abnormal class pools non-normal morphologies. We report macro-F1 to account for imbalance. The abnormal class pools structurally distinct morphologies (LBBB, RBBB, PVC, APB), justifying macro-F1 as the primary metric. Few-shot personalization uses $K\in\{1,5,10\}$ support examples per class. Records without at least $K+1$ minority-class beats are excluded for that shot count.

    Firmware validation targets the CY8CPROTO-063-BLE board, which integrates a PSoC~6 BLE device with a Cortex-M4F application core and 1 MB Flash / 288 KB SRAM~\cite{infineon2020guide}. The replay firmware stores exported weights and capped support/query episodes in flash, executes the frozen backbone and both head rules on the CM4, and logs metrics over UART. DWT cycle counts are collected from the CM4 Release build. The measured HF0 clock during profiling is 100 MHz.

    Table~\ref{tab:memory} reports the model-centric memory budget for the deployable adaptation path. We report this separately from the full replay firmware image, which also contains board-support code, UART/profiling logic, the C runtime, and embedded replay assets. The source code to fully reproduce the work is available in~\url{https://github.com/eduluzufop/TinyML-ECG-Prototype-Adaptation}.

    \begin{table}[!htb]
        \centering
        \caption{Model-centric memory budget of the deployable adaptation path on PSoC~6. Weights are stored as \texttt{float32} constants in flash.}
        \label{tab:memory}
        \begin{tabular}{lcc}
            \toprule
            Item & Flash & SRAM \\
            \midrule
            Backbone weights & 4.9 KB & -- \\
            Head/prototypes & 0.3 KB & 0.3 KB \\
            Peak activation buffer & -- & 6.3 KB \\
            Support buffer, $K=10$ & -- & 15.6 KB \\
            \midrule
            Subtotal & 5.2 KB & 22.2 KB \\
            PSoC~6 budget & 1024 KB & 288 KB \\
            Utilization & 0.5\% & 7.7\% \\
            \bottomrule
        \end{tabular}
    \end{table}

\section{Results}
\label{sec:results}

    Table~\ref{tab:main} summarizes the accuracy and measured embedded costs for the target tiny backbone. Prototype adaptation consistently outperforms the pre-adaptation model and linear SGD. On the full eligible DS2 protocol, prototypes improve macro-F1 from 0.635 to 0.731 at 1-shot, from 0.639 to 0.771 at 5-shot, and from 0.646 to 0.797 at 10-shot. Linear SGD improves over the baseline but remains lower across all shot counts.

    \begin{table}[!htb]
        \centering
        \caption{Accuracy and embedded cost for the target tiny backbone. Device F1 is computed on the capped 18-record 1-shot replay subset. Timing is measured with the Arm Data Watchpoint and Trace (DWT) cycle counter on the PSoC 6 CM4.}
        \label{tab:main}
        \begin{tabular}{lcccc}
            \toprule
            Method & $K$ & DS2 F1 & Device F1 & Adapt. time \\
            \midrule
            No adaptation & -- & 0.664$^\dagger$ & 0.611 & -- \\
            Linear SGD & 1 & 0.671 & 0.682 & 24.49 ms \\
            Prototype & 1 & 0.731 & 0.798 & 22.83 ms \\
            Prototype & 5 & 0.771 & -- & 114.03 ms \\
            Prototype & 10 & 0.797 & -- & 228.04 ms \\
            Normal-only proto & 10 & 0.699 & -- & $\approx$10 ms$^\ast$ \\
            \bottomrule
        \end{tabular}
        \vspace{1mm}
        \footnotesize
        $^\dagger$Pre-adaptation F1 on all 22 DS2 records; shot-specific baselines are 0.635/0.639/0.646 due to eligibility filtering. $^\ast$Host reference for the restricted update.
    \end{table}

\begin{figure}[!htb]
        \centering
        \includegraphics[width=\columnwidth]{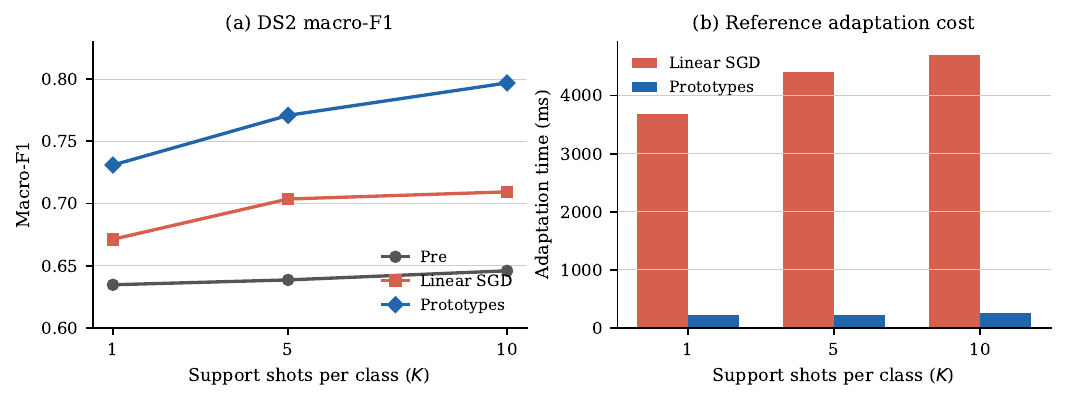}
        \caption{Aggregate DS2 results across shot counts. \emph{Left}: macro-F1 for each method. \emph{Right}: software-side adaptation time from the host reference implementation. The pre-adaptation curve rises with $K$ because the eligibility filter excludes harder records, not due to any adaptation effect.}
        \label{fig:results}
    \end{figure}

    Fig.~\ref{fig:results} shows that the prototype F1 curve rises monotonically from 0.731 to 0.797 as $K$ increases, while SGD gains are smaller, and that prototype calibration is $17{\times}$ faster than SGD at 1-shot in the host reference implementation (216~ms vs.\ 3690~ms; on-device, the gap is negligible at 22.83~ms vs.\ 24.49~ms). The pre-adaptation baseline rises with $K$ because the eligibility filter progressively removes the harder, more class-imbalanced records rather than reflecting any adaptation benefit.

    The aggregate on-device replay benchmark replaces a single-episode sanity check with hardware evidence over 18 DS2 1-shot episodes and 576 query beats. The device-side macro-F1 values are 0.611, 0.682, and 0.798 for pre-adaptation, linear SGD, and prototypes, respectively. Compared with the host reference on the exact same capped subset, the absolute differences are 0.000, 0.007, and 0.000. This confirms that the replay, embedding extraction, and prototype update remain faithful on target hardware.

\begin{figure}[!htb]
        \centering
        \includegraphics[width=\columnwidth]{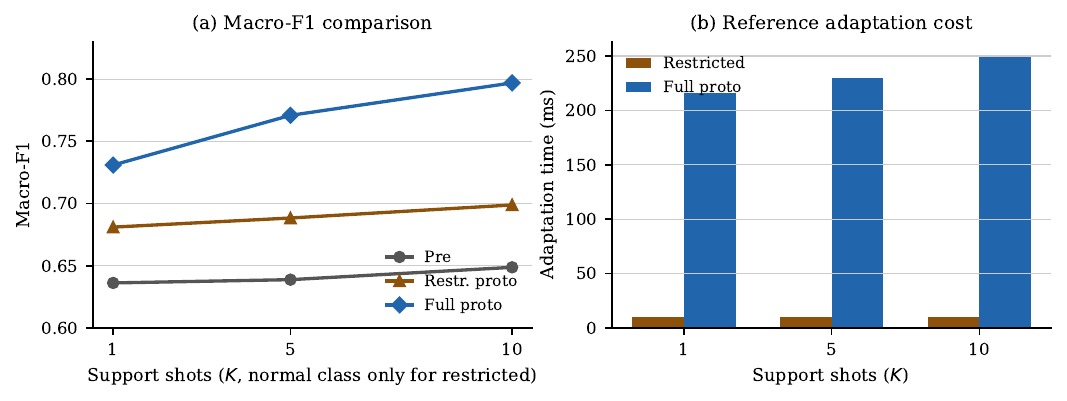}
        \caption{Restricted vs.\ full prototype adaptation on the common eligible DS2 subset. The restricted variant (only normal-beat support from the target patient) provides a consistent positive gain at a fraction of the cost of full prototype adaptation.}
        \label{fig:restricted}
    \end{figure}

    The normal-only calibration mode is weaker than full prototype adaptation but is clinically meaningful. At $K=10$, it reaches 0.699 macro-F1 on the common eligible subset, compared with 0.649 before adaptation and 0.797 with full prototype support. This indicates that patient-specific normal morphology alone can shift the decision geometry in a useful direction, even though it cannot replace labeled abnormal support. Fig.~\ref{fig:restricted} shows that the gain is consistent across all shot counts ($+$0.045--0.050 macro-F1), representing roughly one-third of the full prototype improvement at $K{=}10$ without requiring any arrhythmia annotation from the target patient. This is the intended deployment role of the restricted mode, an initial calibration step based on abundant (normal) sinus beats.

\begin{figure}[!htb]
        \centering
        \includegraphics[width=\columnwidth]{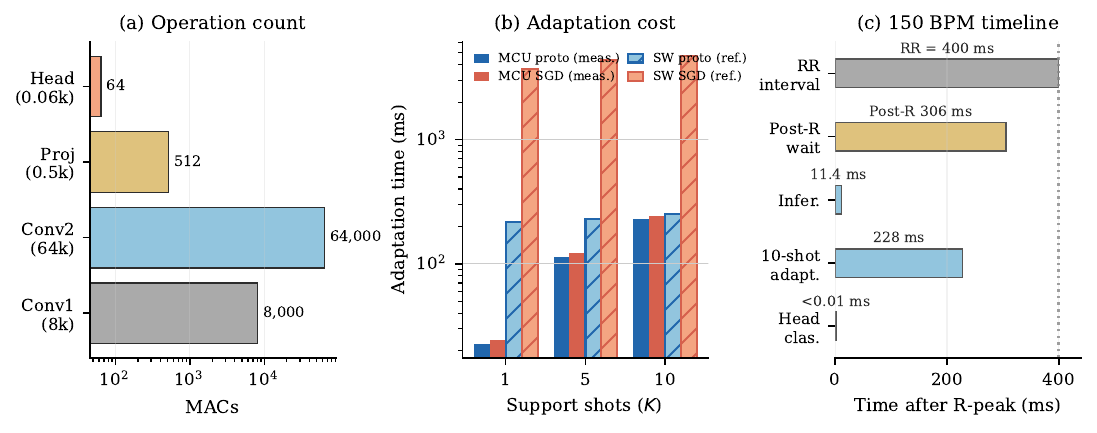}
        \caption{End-to-end calibration time on the PSoC~6 CM4 at 100 MHz vs.\ number of support shots $K$ for prototype and linear SGD adaptation. Both methods scale linearly with $K$ (one backbone forward pass per support beat); the gap between the two curves is due solely to the head-update rule. The 400 ms RR interval at 150 BPM is shown for clinical reference.}
        \label{fig:compute-cost}
    \end{figure}

    Measured DWT profiling reports 1,139,141 cycles per beat for the tiny backbone, corresponding to 11.39 ms at 100 MHz, 2.8\% of the 400 ms beat-to-beat interval (a.k.a RR interval) at 150 BPM, confirming that the device classifies each beat in time even at clinically elevated heart rates. Fig.~\ref{fig:compute-cost} plots end-to-end calibration time on the CM4 as a function of $K$ for both methods. Both curves grow linearly with $K$ because each method requires one backbone forward pass per support beat. The narrow gap between them reflects only the head-update rule (closed-form mean vs.\ 50 SGD iterations). A 10-shot prototype calibration takes 228.04 ms (22,804,485 cycles), and the corresponding SGD calibration takes 244.66 ms, both within the 400 ms RR interval, so calibration is a short initialization event rather than a recurring per-beat cost.

    Table~\ref{tab:scaling} evaluates whether the prototype advantage is preserved under backbone scaling. The tiny backbone is both the primary embedded target and the setting where prototypes most consistently outperform SGD. Larger backbones narrow or reverse this advantage at low shot counts, indicating that the prototype-versus-SGD trade-off is capacity-dependent rather than a universal property of the head-adaptation approach.

    \begin{table}[!htb]
        \centering
        \caption{Scaling study across backbone sizes. DS2 test F1 is reported before personalization. Few-shot columns show the best retuned head-adaptation method.}
        \label{tab:scaling}
        \begin{tabular}{lccccc}
            \toprule
            Variant & Params & MACs & Pre F1 & 1-shot & 10-shot \\
            \midrule
            Tiny & 1,314 & 72.6k & 0.664 & Proto .731 & Proto .797 \\
            Medium & 2,618 & 157.0k & 0.642 & SGD .713 & Proto .764 \\
            Large & 4,930 & 274.2k & 0.651 & SGD .720 & Proto .759 \\
            \bottomrule
        \end{tabular}
    \end{table}

    The consistent prototype advantage on the tiny backbone can be attributed to the geometry of its compressed embedding space. With $D=32$ and only 1,314 parameters, the intra-class variance is tightly bounded, making the closed-form mean a stable representative~\cite{snell2017prototypical}, a condition that more expressive estimators such as task-guided diffusion~\cite{du2023protodiff} or prototype-neighbor hybrids~\cite{jiang2025prototype} are designed to relax, at the cost of auxiliary networks and iterative inference that exceed the MCU budget. Larger backbones produce richer manifolds where the mean prototype becomes a less faithful class summary, shifting the advantage toward SGD at low shot counts (Table~\ref{tab:scaling}).

\section{Discussion, Limitations, and Conclusion}

    For the target tiny backbone, prototype adaptation is preferable to linear SGD because it directly matches the embedded constraint: no convolutional backward pass, no optimizer state, and a closed-form update in a 32-D embedding space. This is particularly attractive for patient-specific ECG calibration, where the support set is extremely small and gradient-based head updates can be ill-conditioned. The result complements prior on-device training work~\cite{lin2022ondevice} and on-device HAR personalization~\cite{craighero2024ondevice} by exploring a point in the design space where adaptation is reduced to nearest-centroid updates rather than training. The per-class independence of this update also underlies the restricted normal-only variant, because $p_0$ and $p_1$ are updated separately, calibration can begin from passively buffered sinus (normal) beats alone (with no arrhythmia annotation required), a property that SGD-based head fine-tuning does not share, since it requires labeled examples from all classes to update the weight matrix meaningfully.

    The scaling study shows that larger backbones do not automatically improve inter-patient macro-F1, and they make the prototype-versus-SGD ranking shot-dependent. Therefore, our claim is not that prototypes dominate all personalization regimes, but that they are especially well matched to the target tiny model and MCU deployment path.

    Since this work focuses on the embedded personalization primitive rather than a complete diagnostic pipeline, beat segmentation and R-peak detection are assumed to be upstream, and the classification task is binary (normal versus abnormal) rather than the full AAMI five-class taxonomy. The aggregate on-device replay uses a flash-capped subset of 18 one-shot DS2 episodes, although it matches the host reference on that subset. The current firmware uses native C \texttt{float32} kernels rather than an int8 deployment with TensorFlow Lite for Microcontrollers (TFLM)~\cite{david2021tflm}.

    In conclusion, this letter presents a prototype-only head adaptation mechanism for TinyML ECG personalization on a PSoC~6 microcontroller. By freezing the 1-D CNN backbone and replacing gradient-based head fine-tuning with closed-form class means, the method improves inter-patient macro-F1 for the target tiny model while remaining feasible within a small measured memory and timing footprint. The restricted normal-only variant further demonstrates that a meaningful personalization gain could be achievable from passively buffered normal beats alone, reducing the need for arrhythmia annotations during initial device deployment. Future work will extend the firmware to int8/TFLM, integrate upstream beat detection, and evaluate the full AAMI five-class setting.

\bibliographystyle{main}  
\bibliography{main}  

\end{document}